\documentclass{article}
\usepackage{subfigure}
\usepackage{authblk}
\usepackage{amsmath}
\usepackage[super]{nth}
\usepackage{setspace}
\usepackage{graphicx}
\usepackage{caption}
\usepackage[T1]{fontenc}

\newcommand{\tabincell}[2]{\begin{tabular}{@{}#1@{}}#2\end{tabular}}

\begin{document}

\title{Monocular Pedestrian Orientation Estimation Based on Deep 2D-3D Feedforward}
\author[]{Chenchen Zhao}
\author[]{Yeqiang Qian}
\author[]{Ming Yang \thanks{Corresponding author: mingyang@sjtu.edu.cn}}
\affil[]{Department of Automation, Shanghai Jiao Tong University, Shanghai, 200240 \authorcr Key Laboratory of System Control and Information Processing, Ministry of Education of China, Shanghai, 200240}
\date{}
\maketitle

\doublespacing

\begin{abstract}
Accurate pedestrian orientation estimation of autonomous driving helps the ego vehicle obtain the intentions of pedestrians in the related environment, which are the base of safety measures such as collision avoidance and prewarning.
However, because of relatively small sizes and high-level deformation of pedestrians, common pedestrian orientation estimation models fail to extract sufficient and comprehensive information from them, thus having their performance restricted, especially monocular ones which fail to obtain depth information of objects and related environment.
In this paper, a novel monocular pedestrian orientation estimation model, called FFNet, is proposed.
Apart from camera captures, the model adds the 2D and 3D dimensions of pedestrians as two other inputs according to the logic relationship between orientation and them.
The 2D and 3D dimensions of pedestrians are determined from the camera captures and further utilized through two feedforward links connected to the orientation estimator.
The feedforward links strengthen the logicality and interpretability of the network structure of the proposed model.
Experiments show that the proposed model has at least 1.72\% AOS increase than most state-of-the-art models after identical training processes.
The model also has competitive results in orientation estimation evaluation on KITTI dataset.
\end{abstract}

\section{Introduction}\label{intro}

Orientation estimation is a significant step of obtaining comprehensive information of 3D objects \cite{p1}.
In autonomous driving, accurate estimation of pedestrian orientation helps the ego vehicle obtain intentions (e.g. details of movements and future trajectories) of pedestrians in the related environment, and improve its self-decisions according to the intentions \cite{p16}.
Such self-decisions include basic policy makings such as path planning and steering control, and safety measures such as collision avoidance and accident prewarning \cite{p18, p19}.

Orientation estimation is also one of the most difficult tasks among all steps of pedestrian information extraction, especially in monocular solutions, with reasons as follows:

\begin{itemize}
\item
More complex appearances \cite{p17} .
Appearances of pedestrians are usually composed of multiple colors and patterns on garments and ornaments.
This makes it easier for monocular solutions to confuse them with the background environment.
Since orientation and patterns are largely irrelevant, monocular solutions face a harder problem of finding orientation-related characteristics.
\item
Smaller sizes.
Pedestrians have smaller height and far smaller width and length than other traffic objects in the environment (e.g. vehicles), making them more difficult to be detected.
The subsequent orientation estimation process is certainly more challenging, since smaller sizes reduce the differences between different orientations of pedestrians.
\item
Higher-level deformation \cite{p17}.
Unlike other traffic objects, pedestrians are deformable, and have quite different shapes and sizes under different states and gestures.
This indicates that a variety of approaches, such as 3D modeling, are not suitable for pedestrian orientation estimation.
Moreover, deformation indicates that the same orientation may correspond to a variety of shape characteristics, which is highly confusing.
\end{itemize}

From the analyses above, it is obvious that accurate pedestrian orientation estimation requires very deep and diverse information of pedestrians, as well as related high-performance information processors.
This is a huge challenge for orientation estimation solutions, especially monocular ones which lack depth information of objects and the related environment.
This results in performance restrictions of the solutions.
Therefore, orientation estimation solutions need information enhancements.

For monocular solutions in which camera captures are the only input data, information enhancement relies largely on internal information mining.
Most state-of-the-art solutions have the same feature extraction and recognition process dealing with both vehicles and pedestrians, thus neglecting the requirement for more information of pedestrian orientation estimation.
Therefore, almost all state-of-the-art monocular solutions have relatively lower performance and huger growth potential in pedestrians.
The official KITTI Object Detection and Orientation Estimation Evaluation benchmark \cite{p2} shows that general precision of pedestrian orientation estimation (represented as average orientation similarity, AOS) is far less than that of cars and cyclists.
Determination of supplementary information and its mining process are urgently needed.

\begin{figure}
\centering
\includegraphics[height=4cm]{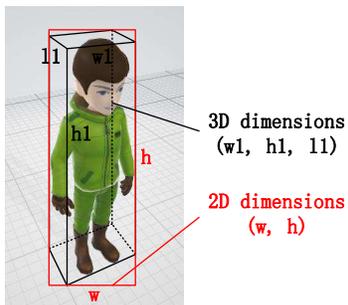}
\caption{Illustration of 2D and 3D dimensions of an exemplary pedestrian.}
\label{pedestrian_2D_3D_illustration}
\end{figure}

Logic relationship is one of the kinds of information that is important but most likely to be neglected.
It lies between inputs and outputs of the solution model, and between outputs if the model has multiple variables to predict.
Since orientation is the only desired variable of the problem, there is no logic relationship without the introduction of another output.
Therefore, in this paper, we introduce the 2D and 3D dimensions of pedestrians and analyze the logic relationship between orientation and them.
The 2D dimensions of the pedestrian are the width and height information of his/her 2D bounding box in the image plane, while the 3D dimensions of the pedestrian are the width, height and length information of his/her 3D bounding box in the camera 3D space.
Illustration of the pedestrian's 2D and 3D dimensions is shown in Fig. \ref{pedestrian_2D_3D_illustration}.

Based on it, a novel pedestrian orientation estimation model, called FFNet, is proposed.
Apart from the camera capture, the model further uses the introduced characteristics (i.e. 2D and 3D dimensions of pedestrians) as two other inputs to make orientation estimations.
In the model, 2D dimensions of the pedestrian are from 2D object detection models, while 3D dimensions of the pedestrian are predictions of the ego model.
To further utilize the two characteristics, two feedforward links are established respectively from 2D and 3D dimensions outputs to the orientation estimator of the model.
Intermediates of the model (i.e. 2D and 3D dimensions of pedestrians) all have physical significance, enhancing the interpretability of the model.
Meanwhile, the feedforward links strengthen the logicality of the model.
Experiments show that the proposed model with 2D-3D feedforward outperforms common state-of-the-art models by 1.72\% AOS with identical training processes.
The proposed model also gets ranked \nth{9} in the official KITTI Object Detection and Orientation Estimation Evaluation benchmark.
Our contributions are summarized as follows:

\begin{itemize}
\item
We propose a novel pedestrian orientation estimation model which has 2D and 3D dimensions as additional inputs compared with other models to make orientation estimations.
The proposed orientation estimation model takes advantage of the relationship between pedestrian's orientation and its 2D and 3D dimensions related to the image plane and camera 3D space respectively, and has outstanding performance compared with most state-of-the-art models.
\item
From the proposed model, we show a way of exploring intermediate properties of a specific problem, followed by another approach (i.e. feedforward links) of exploiting them.
Based on the intermediates and feedforward links, we propose an approach to enhance the logicality and interpretability of neural network models, by utilizing intermediates and establishing feedforward links according to the logic relationship between them, which can be applied to a variety of solution models with multiple predicted variables.
\end{itemize}

The paper is organized as follows:
we discuss the related work in Section \ref{relatedwork}.
In Section \ref{model}, we introduce the proposed model and analyze the logic relationship and feedforward links in it.
We conduct experiments to compare the proposed model with common state-of-the-art ones and make further discussions in Section \ref{experiments}.
We conclude this paper in Section \ref{conclusion}.

\section{Related Work}\label{relatedwork}

There are many Faster R-CNN \cite{p9} based monocular 3D pedestrian detection and orientation estimation solutions having accurate predictions and competitive performance.
Among them, many solution models have competitive results in the KITTI Object Detection \cite{p10, p11, p12, p13, p14, p15} and Orientation Estimation \cite{p3, p4, p5, p6} Evaluation benchmark.
According to the aspects of improvements they focus on, they can be categorized as follows:

\begin{itemize}
\item
Improvements on predictions of the model.
Mono3D \cite{p3} shows an approach of generating bounding boxes in 3D space and projecting them to the image plane.
The authors further proposed several metrics to judge the accurateness of the boxes and make adjustments.
In MonoPSR \cite{p5}, the authors proposed a model that predicts object-centered point cloud information to help determine objects' 3D locations and localizations, along with a novel loss function related to it.
These solutions often have multiple outputs and complex loss functions and metrics related to them.
\item
Improvements on the network structure of the model.
SubCNN \cite{p4} introduces the feature extrapolating layer and the subcategory conv layer to make use of subcategory information in proposal generating.
In Shift R-CNN \cite{p6}, a ShiftNet structure is used for refining 3D object translation predictions.
The ShiftNet takes both 3D object translation predictions and pre-stage 3D dimensions and orientation as inputs, and combines the information to reduce noise disturbance in images.
These solutions may only output predictions of a few basic variables.
However, their related models have complex structures and multiple branches within.
\end{itemize}

All solution models above are from the KITTI Object Detection and Orientation Evaluation benchmark.
They have one shared essence:
mining for extra information for refining or assessing predictions.
However, all of the solutions above focus on information from external sources (i.e. data inputted to the model) rather than internal sources (i.e. intermediates outputted by the model which can be further utilized by itself).
Similarly, external information sources receive much more attention than internal sources.
While a variety of external-information-enhancing algorithms significantly improve performance of neural networks (e.g. pyramid structure for multi-scale detection \cite{p27} and local feature extraction \cite{p28} for typical information-insufficient cases like dimension reduction), mining of internal information is seldom researched.

Besides, they have a common drawback:
using the same tactics to vehicles and pedestrians.
A model that can handle deformable objects with large scale variations is supposed to be more advantageous in pedestrian orientation estimation.
With the task becoming harder, the solution model is required to have the ability to extract more and deeper features in images.
Apart from solution models in the KITTI benchmark, there are a variety of pedestrian detection models focusing on the unique characteristics of pedestrians, based on a variety of sensor inputs not limited to monocular cameras.
In \cite{p20}, multiple parts of pedestrians are respectively recorded and learned by different detectors, which are combined by a decision tree, to solve occlusion problems.
In \cite{p21, p22, p23}, thermal images are studied and applied in nighttime pedestrian detection.
Thermal images take advantage of temperature features of human, and have positive effects on differentiating pedestrians from the background environment.
In \cite{p29}, illumination conditions are further considered to make better combination and use of color and thermal information of images.
In \cite{p30} and \cite{p31}, features of human attributes and 3D poses, as well as their relationships with image features, are respectively studied and combined with visual features of images for related recognition tasks.
In \cite{p24}, Oriented Spatial Transformer Network (OSTN) is proposed to rectify distorted pedestrian features in a detector-friendly manner in pedestrian detection with fish-eye cameras.

None of the solutions above have innovation in orientation estimation.
Most orientation estimation models have one more branch compared with object detection models, used to output orientation predictions.
However, there are tough cases in which extra branches are not qualified for accurate orientation estimations.
Such cases require more complex network structures to deal with.
In this paper, the proposed model focuses on the logic relationship between orientation and other characteristics of pedestrians, which is neglected by all models above.

\section{Pedestrian orientation estimation model based on deep 2D-3D feedforward}\label{model}
\subsection{Logic relationship in pedestrian orientation estimation}\label{logic}

Pedestrian orientation estimation requires more information than vehicle orientation estimation.
For monocular models, utilization of external information can hardly be improved.
Therefore, mining of internal information is significant in this case.

Logic relationship is the most significant internal information.
Use of logic relationship in neural networks has another powerful advantage: avoiding predictions not according with ground truth (i.e. illogical in external terms), or contradictory (i.e. illogical in internal terms).
In deep learning solutions, the logic relationship between inputs and outputs is the base of network training, with related information stored in weights in network layers.
In this paper, we pay main attention to the logic relationship between outputs.
In the proposed model, we introduce 2D and 3D dimensions of pedestrians to the network, to construct logic links to orientation estimation from them.

\begin{figure}
\centering
\includegraphics[height=3cm]{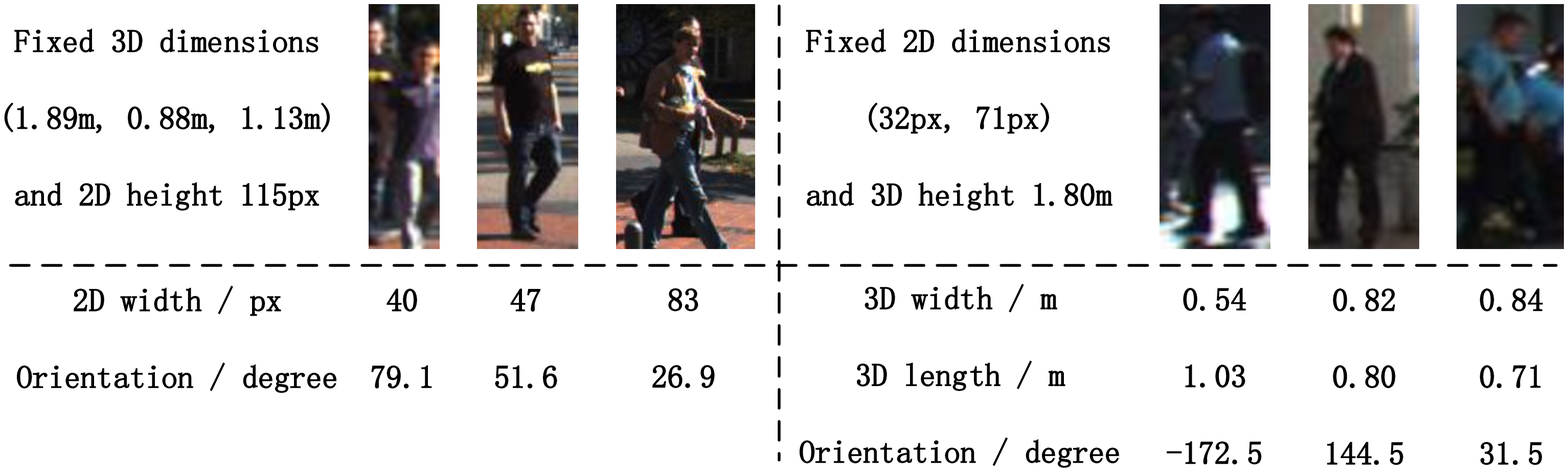}
\caption{Logic relationship between the 2D-3D dimensions and orientation of exemplary pedestrians.}
\label{3D_2D_orientations}
\end{figure}

2D and 3D dimensions, and orientation angles of exemplary pedestrians from the KITTI Object Detection dataset are listed in a control-variate manner in Fig. \ref{3D_2D_orientations}.
To eliminate the influences of scaling on the final results, selected pedestrians' 2D (3D) height is scaled equal when 3D (2D) dimensions are fixed in the control variate method.
Figure \ref{3D_2D_orientations} shows the monotonic influences of pedestrians' 2D and 3D dimensions on its orientation angle respectively.
Assumed that distortion in camera captures has already been rectified, while the camera and pedestrians are at the same height, the width span of pedestrians in camera X-axis is as Equation \ref{eq1}.

\begin{equation}
\overline{w_1}=
\begin{cases}
w_1\sin\theta+l_1\cos\theta, &\text{if $0\le\theta<\frac{\pi}{2}$}\\
w_1\sin\theta-l_1\cos\theta, &\text{if $\frac{\pi}{2}\le\theta<\pi$}\\
l_1\cos\theta-w_1\sin\theta, &\text{if $-\frac{\pi}{2}\le\theta<0$}\\
-l_1\cos\theta-w_1\sin\theta, &\text{if $-\pi\le\theta<-\frac{\pi}{2}$}
\end{cases}
\label{eq1}
\end{equation}

Values of 2D dimensions ($h$, $w$), 3D height $h_1$ and width span $\overline{w_1}$ satisfy Equation \ref{eq2}.

\begin{equation}
\frac{h}{w}=\frac{h_1}{\overline{w_1}}
\label{eq2}
\end{equation}

Corresponding to Fig. \ref{3D_2D_orientations}, Equation \ref{eq1} and \ref{eq2} represent the logic relationship in mathematic terms:
with $\theta$ changing in the range $(-\pi, \pi)$, $\overline{w_1}$ changes with a relatively large amplitude, which can be reflected in the ratio of $h$ to $w$.
This can be used in pedestrian cases, especially walking pedestrians which have huge differences between $w_1$ and $l_1$.

Since object orientation is desired, 2D and 3D dimensions are at the beginning of the logic links, while orientation is at the end.
Therefore, the logic links are established from 2D and 3D dimensions to orientation in the proposed model.

\subsection{2D-3D feedforward based pedestrian orientation estimation model}\label{sectionnetwork}

The proposed FFNet model is based on the implementation of the logic links proposed above.

For the implementation of a logic link, the most direct approach is to make those variables at the beginning of it as inputs, and those variables at the end of it as outputs, to make the value of the former variable influence prediction of the latter one.
Therefore, in the network of the proposed model, there are layers connected to orientation output, with values of 2D and 3D dimensions as inputs.
These layers with 2D and 3D dimensions inputs are respectively combined and represented by feedforward links in the proposed model.

The feedforward mechanism in neural networks has many successful implementations in object detection.
In Shift R-CNN \cite{p6}, a feedforward link lies between the predicted characteristics of the object and the 3D translation predictor, to take the object's 2D and 3D characteristics as inputs for its 3D translation regression.
In Mask Scoring R-CNN \cite{p7}, the mask IoU prediction process takes the predicted mask as input as well as RoI features to achieve a more accurate mask IoU prediction result.
In the field of click feature prediction in image recognition, feedforward is used for label generation in semi-supervised learning \cite{p26} and supplementary information (i.e. hypergraph manifold) offering \cite{p32}.
However, all solution models above use feedforward for regression or error checking instead of direct prediction, and none of them pays attention to the logic relationship between the predicted characteristics of objects.
In this paper, the proposed model implements feedforward in a direct one-step prediction process.

\begin{figure}
\centering
\includegraphics[height=3cm]{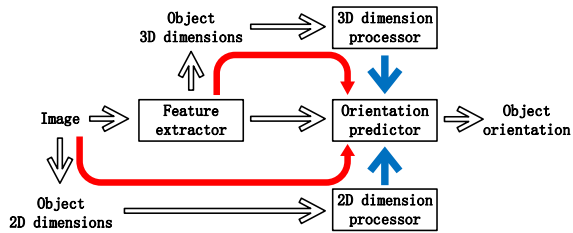}
\caption{The overall orientation estimation process of the proposed model.}
\label{process}
\end{figure}

The overall orientation estimation process of the proposed model is shown in Fig. \ref{process}, in which the feedforward links are marked red, and the logic links to orientation estimation respectively from 2D and 3D dimensions of pedestrians are marked blue.
Input of the 3D feedforward link is provided by the feature extractor in the network model.
Therefore, the network model makes 3D dimensions regression before making orientation estimation, and the 3D dimensions result is used as input to the 3D feedforward link.
In contrast, input of the 2D feedforward link is considered as external information, and is provided by an external 2D object detection model.
In this paper, the 2D object detection model \cite{p14} is adopted for providing the 2D dimensions information.

\begin{figure}
\centering
\includegraphics[height=3cm]{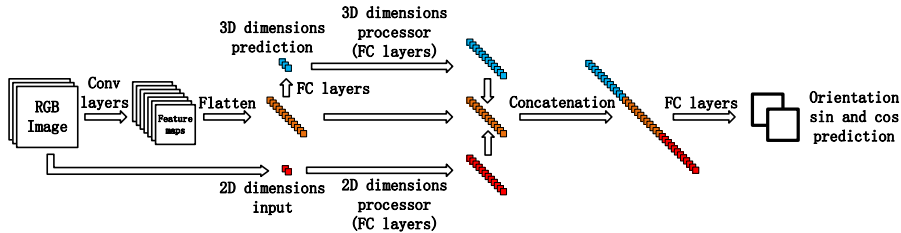}
\caption{Network structure of the proposed model.}
\label{network}
\end{figure}

Based on the feedforward links above, the proposed model has the network structure shown in Fig. \ref{network}.
The network adopts the VGG16 \cite{p25} frame, with added 2D and 3D dimensions processors and a related concatenation process.
The 2D and 3D dimensions processors are constructed by 3 fully-connected layers, with the number of neurons 2 (2D dimensions processor) / 3 (3D dimensions processor), 512 and 2048 respectively.
Outputs of the two processors are then directly concatenated to the flattened feature maps outputted by the conv layers.

Direct concatenation is widely adopted in the implementation of feedforward.
In Shift R-CNN, prediction of 3D translation, 2D bounding box information, 3D dimensions and orientation are all combined and inputted to the 3D translation refinement network.
In Mask Scoring R-CNN, RoI feature maps and the predicted mask are stacked to form an ensemble and fed into the mask scoring network.
Since outputs of 2D and 3D dimensions processors have no planarity other than outputs of the two models above, the concatenation process connects elements of outputs to form a larger-scale fully-connected layer.
The fully-connected layer is then connected to 2 fully-connected layers with the number of neurons 512 and 2 respectively.
The last fully-connected layer outputs $\sin$ and $\cos$ values of the orientation prediction.

Note that in two network structures lie the way of extracting and exploiting the 2D and 3D dimensions information the network model adopts: the fully-connected layers in the feedforward links and after the concatenation process.
The extraction and exploitation of the information are not label-guided, since it is encouraged that the network model takes consideration of the related environment and possible ground truths, which cannot be labeled.
The effectiveness of the model is reflected by the accuracy of its orientation output, and is improved through the training process.

In the 3D feedforward link, the back-propagation process starts from orientation prediction, and continues until it reaches the first conv layer.
This makes the errors of orientation estimation influence the feature extraction and 3D dimensions prediction processes above.
To avoid this, gradient truncation is used at the 3D dimensions output (i.e. preventing the gradient from propagating towards the 3D dimensions prediction process), to eliminate the problems raised above.

\subsection{Definitions of loss functions}\label{loss}

The loss function is as Equation \ref{eq3}.

\begin{equation}
    loss=loss_{dimensions}+loss_{orientation}
\label{eq3}
\end{equation}

Dimensions loss adopts the L2-loss form, while orientation loss has the form as Equation \ref{eq4}.

\begin{equation}
\begin{split}
&loss_{orientation}=\sum_{i=1}^{bin}loss_i\\
loss_i=1 &-\sin\overline{orientation_i}\times\sin orientation_i\\
         &-\cos\overline{orientation_i}\times\cos orientation_i
\end{split}
\label{eq4}
\end{equation}

In Equation \ref{eq4}, $orientation_i$ and $\overline{orientation_i}$ are respectively the estimated and labeled orientation value.
The orientation loss is positive if orientation output does not equal its truth value, and has a positive correlation with the difference between them.

Meanwhile, determination of the estimated orientation angle has the form as Equation \ref{eq5}.

\begin{equation}
\begin{split}
orientation_i &=\arctan\frac{SinValue_i}{CosValue_i}\\
              &+
\begin{cases}
0, &\text{if $CosValue_i>0$}\\
\pi, &\text{if $CosValue_i<0$ \& $SinValue_i>0$}\\
-\pi, &\text{if $CosValue_i<0$ \& $SinValue_i<0$}
\end{cases}
\end{split}
\label{eq5}
\end{equation}

\begin{figure}
\centering
\includegraphics[height=4cm]{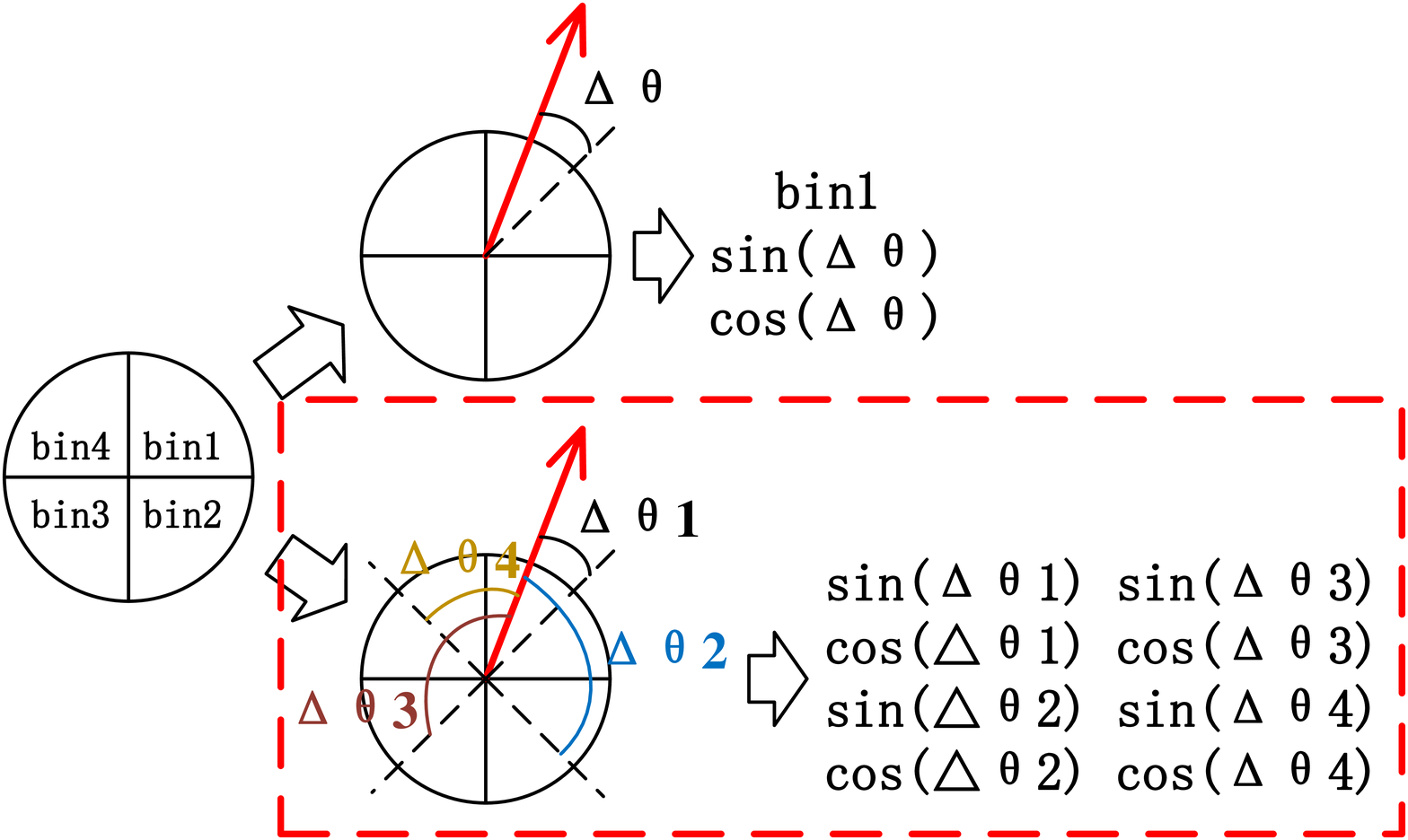}
\caption{The proposed orientation estimation algorithm based on the MultiBin algorithm.}
\label{multibin}
\end{figure}

In Equation \ref{eq5}, $SinValue_i$ and $CosValue_i$ are respectively $\sin$ and $\cos$ output of the network.
In some ways, the orientation loss is like the MultiBin orientation loss proposed in the Deep3DBox model \cite{p8}.
The MultiBin orientation estimation algorithm performs two steps.
First, it divides the orientation space into multiple bins and predicts the bin the orientation angle stays in.
Specifically, if the orientation angle stays near the edge of two bins, MultiBin considers it `in both bins', and assigns classification confidence 0.5 to each bin.
Second, it predicts the residual rotation between the orientation angle and the center of the predicted bin.
The two-step process of the MultiBin algorithm is shown in the upper branch in Fig. \ref{multibin}.

Although the MultiBin algorithm simplifies the orientation estimation problem and turns it into semi-classification-semi-regression, it has two main drawbacks:

\begin{itemize}
\item
MultiBin loss neglects the differences between angles near the edges and angles with little classification confidence, since the algorithm outputs maximum classification confidence 0.5 for angles near the edges.
The confusion in classification confidence directly influences confidence of orientation estimation in the Deep3DBox model.
\item
MultiBin relies largely on the classification result at the first stage, thus introducing another type of error to the orientation estimation process.
In the case shown in Fig. \ref{multibin}, misclassification results in $90^\circ$ orientation error in average.
\end{itemize}

In order to mitigate the two problems of MultiBin, the proposed orientation loss abandons classification, but adopts the multi-bin mechanism, and takes consideration of prediction results from all bins.
The proposed orientation estimation algorithm is shown in the lower branch in Fig. \ref{multibin}.
In this way, the first problem can be eliminated.
In the meantime, multiple prediction results act as voters to exclude false outputs among them, with excluding algorithm presented below.

In ideal conditions, after offset adjustments, orientation outputs from all bins should be equal.
For orientation output from each bin, if it satisfies both of the following 2 conditions, it will be excluded from orientation loss calculating.

\begin{itemize}
\item
Differences between it and outputs from other bins are all greater than a predefined threshold.
\item
Differences between outputs from other bins are all smaller than the threshold.
\end{itemize}

\section{Experiments, results, and discussions}\label{experiments}

\subsection{Experiment setup}\label{setup}

The proposed model is tested and compared in two aspects.
First, we construct a plain model, which is the same as the proposed model but without logic relationship and 2D-3D feedforward, and compare the performance of them.
Second, we test the proposed model on the KITTI Object Detection and Orientation Estimation Evaluation benchmark and compare it with other state-of-the-art models.

In order to investigate the effect of feedforward links on feature combination, we construct a new loss function according to Equation \ref{eq1} and \ref{eq2}:

\begin{equation}
\begin{split}
\overline{w_1}_{prediction}={w_1}_{prediction}\times\left|\sin\theta\right|+{l_1}_{prediction}\times\left|\cos\theta\right|\\
\overline{w_1}_{prediction}^1=w_1\times\left|\sin\theta_{prediction}\right|+l_1\times\left|\cos\theta_{prediction}\right|
\end{split}
\label{eq6}
\end{equation}

\begin{equation}
\begin{split}
loss\_consistency_{dimensions}=h\times\overline{w_1}_{prediction}-w\times h_1\\
loss\_consistency_{orientation}=h\times\overline{w_1}_{prediction}^1-w\times h_1
\end{split}
\label{eq7}
\end{equation}

The two consistency losses are added to the total loss (i.e. Equation \ref{eq3}) with weight 0.01 each.
The consistency loss combines 2D and 3D dimensions with orientation prediction in the loss function perspective, without intervention of feedforward links.

For comparisons in the first aspect, there are four models involved: the proposed model and the plain model, respectively with and without consistency loss.
The four models are trained and validated through identical training and validation processes with the same dataset.
Since the KITTI benchmark is used for testing the proposed model, the related KITTI dataset is used for training and validation.
5-fold cross validation is conducted, with 6000 images as training data and 1481 images as validation data.
Parameters and environment of the training and validation processes are listed in TABLE \ref{param}.
For comparisons in the second aspect, the well-trained model in the first aspect is tested in the KITTI benchmark, with results and rankings compared with other models.

\begin{table}[]
\centering
\caption{Parameters and environment of the training and validation process.}
\begin{tabular}{cc}
\hline\hline
Parameters / Environment & Value / Choice\\
\hline
\tabincell{c}{Number of bins in the\\multi-bin mechanism} & 4\\
Hardware environment & NVIDIA GeForce GTX 1080Ti\\
Batch size & 32 (training) / 1 (validation)\\
Number of batches (training) & 22500\\
Learning rate (training) & \tabincell{c}{$10^{-3}$ (first $6\times 10^3$ batches)\\$10^{-4}$ (last $1.65\times 10^4$ batches)}\\
\hline\hline
\end{tabular}
\label{param}
\end{table}

In training and validation, learning speed, minimum loss value, inference accuracy, runtime resource occupancy and runtime speed are all concerned metrics in the comparisons.

\subsection{Experiment results with the plain models}\label{resultplain}

\begin{figure}
\centering
\includegraphics[height=4cm]{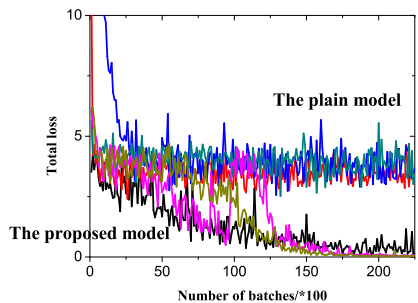}
\caption{Loss convergence process of the proposed model and the plain model in training (without consistency loss). 3 of the 5-fold cross validation processes are shown.}
\label{totalloss}
\end{figure}

In training, given the same loss function, the proposed and the plain model have the loss convergence processes shown in Fig. \ref{totalloss}.
While the two models have similar speed of loss convergence at the first few batches, the plain model reaches the minimum loss value roughly at the \nth{5000} batch while the proposed model still has a stable loss decreasing trend.
The proposed model reaches the minimum loss value roughly at the \nth{17500} batch.
The proposed model has the minimum loss value of only 0.13, while that of the plain model is 3.4.
Therefore, logic links and 2D-3D feedforward help to enhance the training performance of the network model.

\begin{figure}
\centering
\subfigure[]{\includegraphics[height=4.3cm]{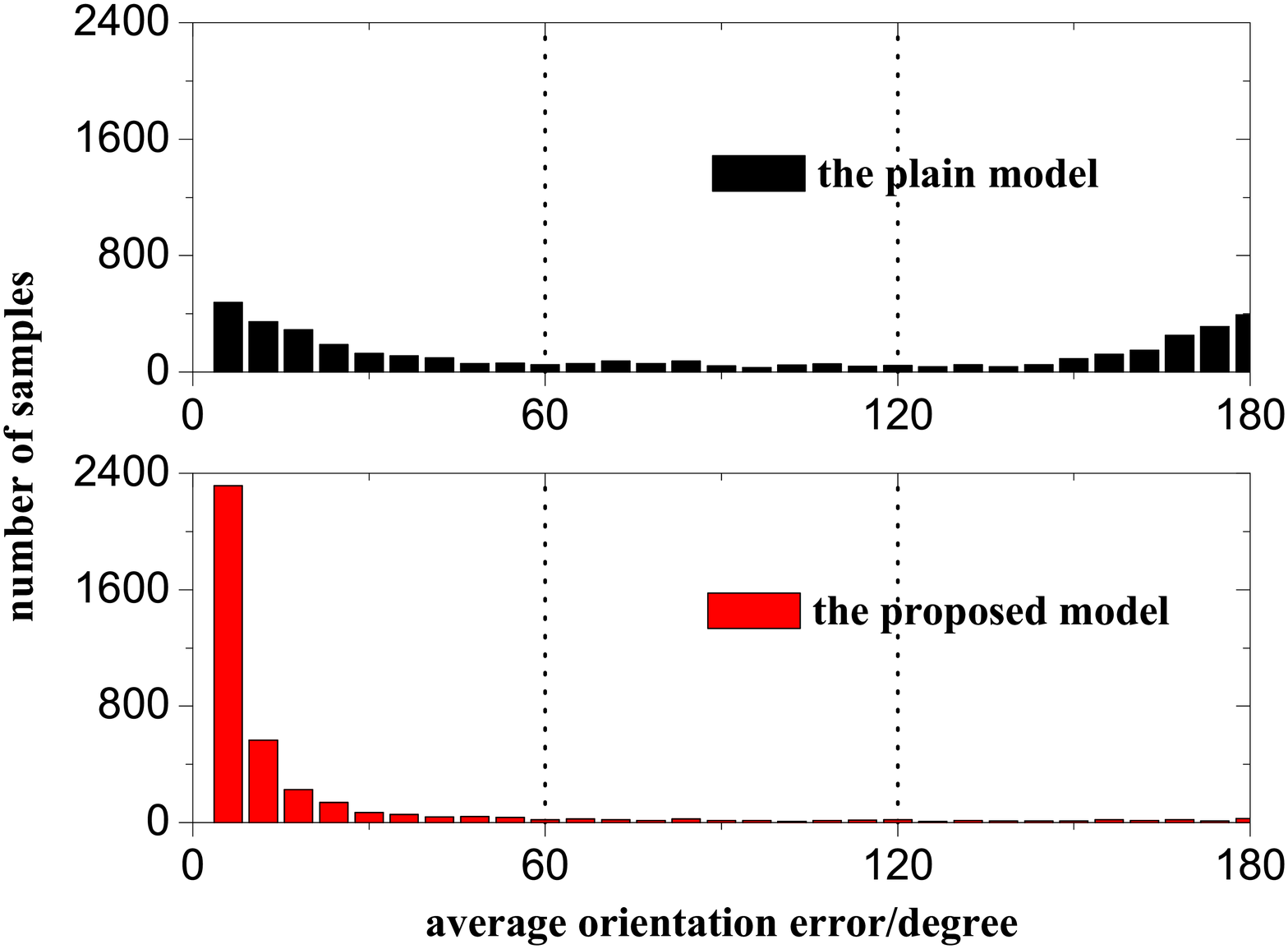}}
\subfigure[]{\includegraphics[height=4.3cm]{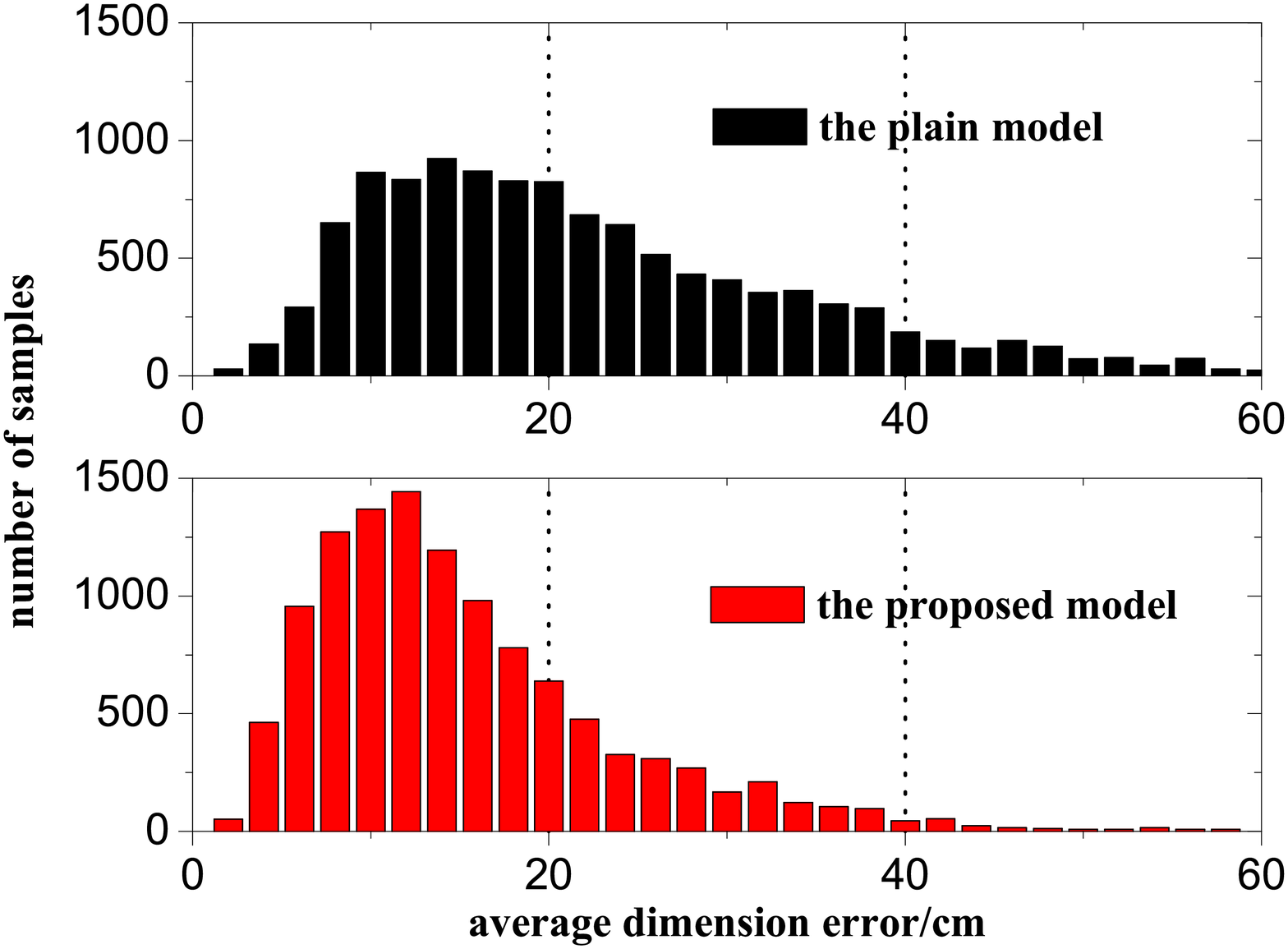}}
\caption{Orientation and 3D dimensions error distributions of the proposed and the plain model on all validation data samples.}
\label{errordis}
\end{figure}

In validation, given the same loss function, orientation error distributions of the proposed and the plain model on all validation data samples are shown in Fig. \ref{errordis}(a), while 3D dimensions error distributions are shown in Fig. \ref{errordis}(b).
The proposed model has orientation estimation error massed in 0-$30^\circ$, and has far better performance of recognizing the main axes of pedestrians, proven by its much fewer samples with errors around $180^\circ$.
Therefore, logic links and 2D-3D feedforward have positive effects on the network model in validation.
This is in accordance with the better training results of the proposed model shown above.

\begin{figure}
\centering
\includegraphics[height=4cm]{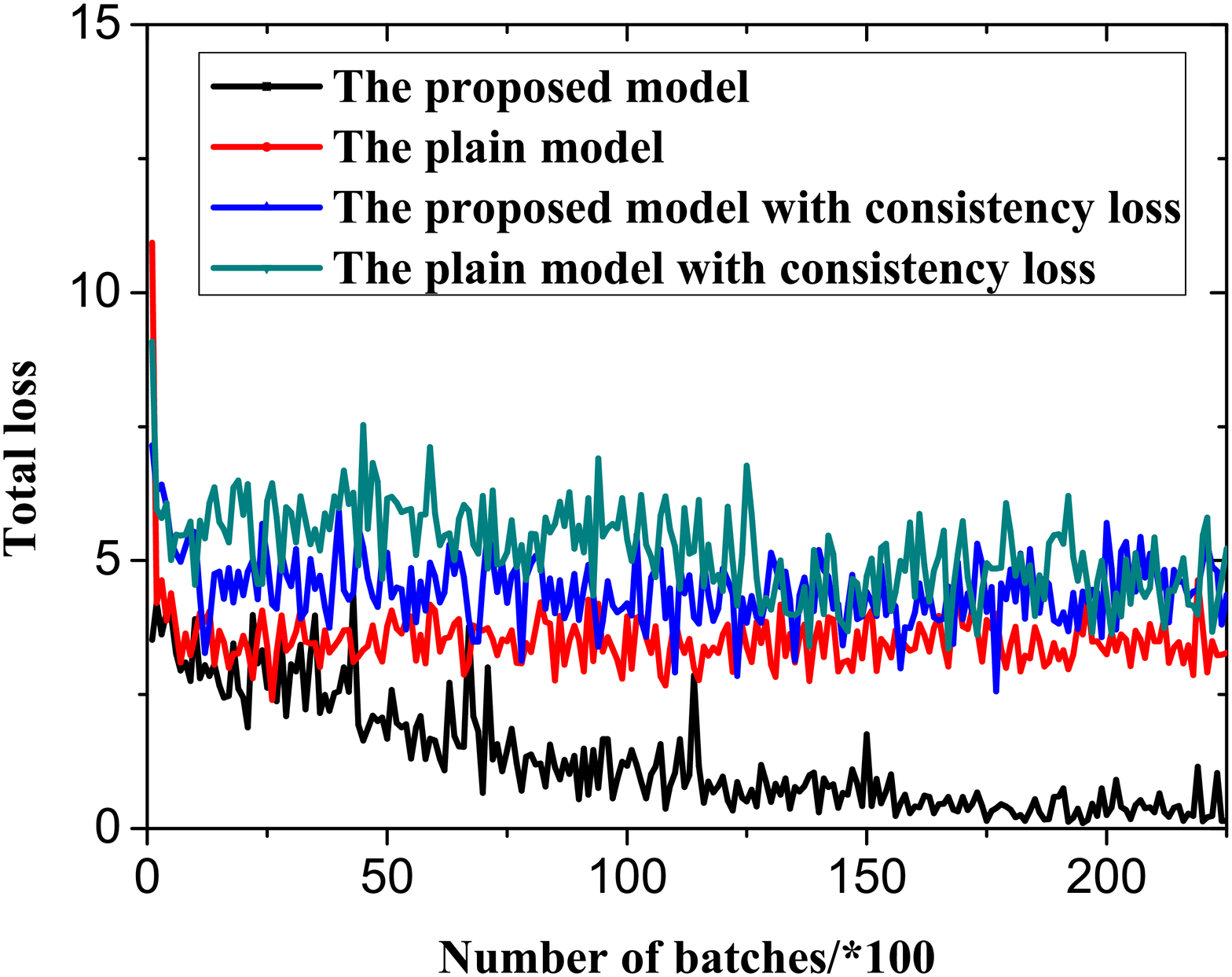}
\caption{Loss convergence process of the proposed model and the plain model in training (with consistency loss).}
\label{losscon}
\end{figure}

With consideration of consistency loss, the proposed and the plain model have the loss convergence processes shown in Fig. \ref{losscon}.
From Fig. \ref{losscon}, it is obvious that consistency loss reduces accuracy of the proposed and the plain model.
For the plain model, this is partly because the network focuses more on converging consistency losses than the parts in Equation \ref{eq3}.
Since consistency losses do not reflect definite accuracy of dimensions regression and orientation estimation, the model is worse at accuracy if with consistency loss.
For the proposed model, another important reason is that consistency losses are not compatible with the approaches the network adopts for extraction and exploitation of 2D and 3D information, which are analyzed in Section \ref{sectionnetwork}.
Therefore, it is wise to let the network learn its way of dealing with 2D and 3D information.
Note that consistency losses are calculated in training, but not calculated in plotting Fig. \ref{losscon}, to make better horizontal comparisons between the models with and without consistency loss.

\subsection{Comparisons of time and resource consumption between the proposed and the plain model}\label{timememory}

In training, the proposed model takes 7.6s and roughly 4500MB runtime memory to learn from 100 batches of data, while those of the plain model are 7.1s and 4400MB.
This is because the added 2D-3D dimensions processors, related feedforward links and the concatenation layer require extra processing time and resources.
However, slight differences in processing time and runtime memory between the two models can be neglected in applications.

In validation, considering the time of reading and preprocessing data, the proposed model takes 22.9s and roughly 3900MB runtime memory to complete the inference process on all validation data (3814 samples), while those of the plain model are 21.8s and 3900MB.
The cause of the differences is the same as that in training, and can also be neglected in applications.

\subsection{Test results in the KITTI benchmark with state-of-the-art models}\label{kitti}

The proposed model is further tested in the KITTI Object Detection and Orientation Estimation Evaluation benchmark. Test results as well as that of other state-of-the-art models \cite{p3, p4, p5, p6} are shown in TABLE \ref{kittiresult}.

\begin{table}[]
\centering
\caption{Benchmark results of the proposed model, the plain model, and other state-of-the-art models.}
\begin{tabular}{cccccc}
\hline\hline
Model & Moderate & Easy & Hard & Runtime & Ranking\\
\hline
\textbf{FFNet (proposed)} & \textbf{59.17\%} & \textbf{69.17\%} & \textbf{54.95\%} & \textbf{0.22s} & \textbf{\nth{2}}\\
Mono3D \cite{p3} & 58.12\% & 68.58\% & 54.94\% & 4.2s & \nth{3}\\
SubCNN \cite{p4} & 66.28\% & 78.33\% & 61.37\% & 2s & \nth{1}\\
MonoPSR \cite{p5} & 56.30\% & 70.56\% & 49.84\% & 0.2s & \nth{4}\\
Shift R-CNN \cite{p6} & 48.81\% & 65.39\% & 41.05\% & 0.25s & \nth{5}\\
Plain & 38.92\% & 46.36\% & 35.63\% & 0.21s & \nth{6}\\
\hline\hline
\end{tabular}
\label{kittiresult}
\end{table}

The proposed model has competitive results in the KITTI benchmark, ranked \nth{2} among common state-of-the-art models and \nth{9} on the leaderboard.
This proves the effectiveness of the logic links and 2D-3D feedforward.

\begin{figure}
\centering
\includegraphics[height=4cm]{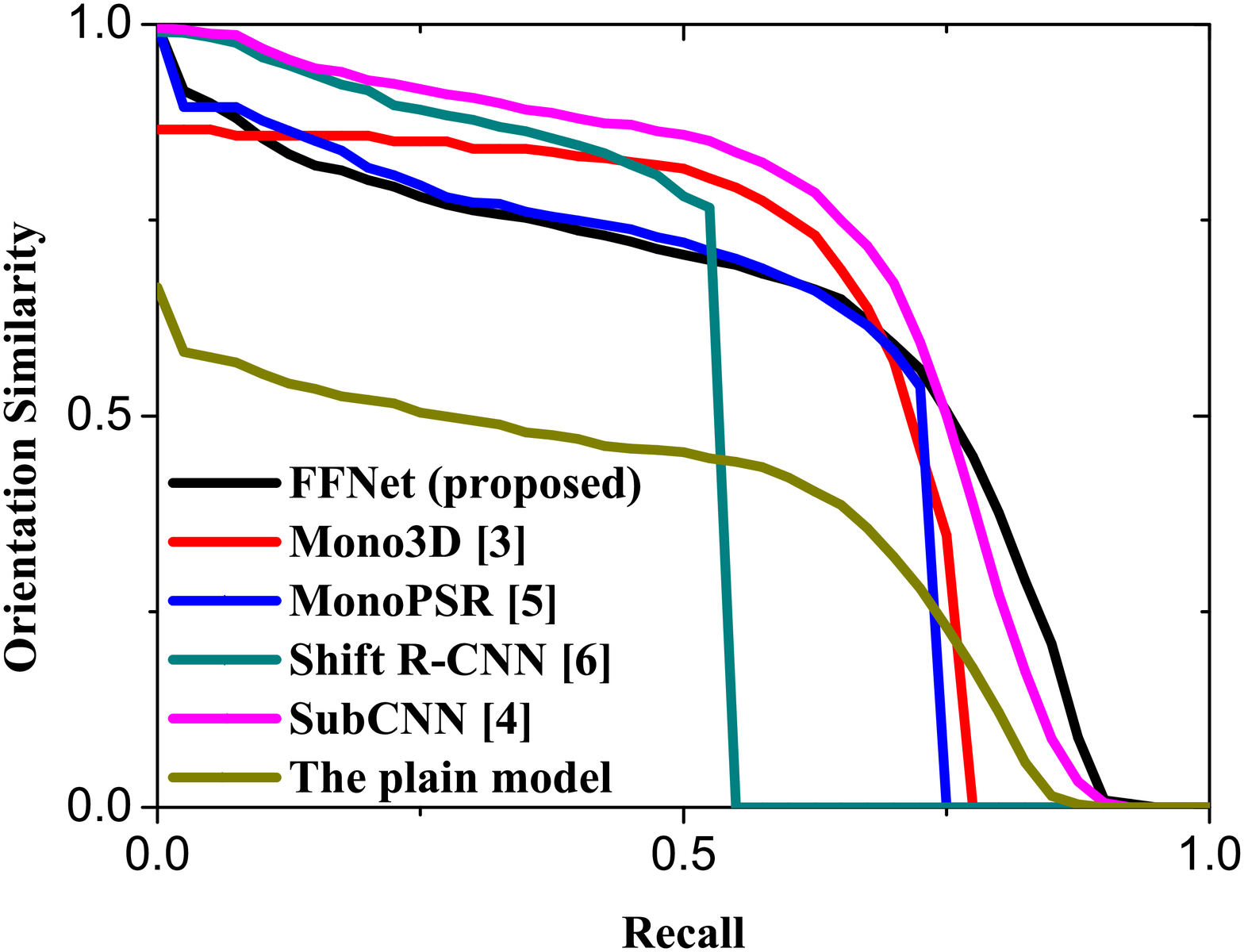}
\caption{OS-R curves of the proposed model, the plain model, and other state-of-the-art models.}
\label{osr}
\end{figure}

Corresponding to the results shown in TABLE \ref{kittiresult}, OS-R (Orientation Similarity \cite{p2} - Recall) curves of the proposed model, the plain model, and other state-of-the-art models are shown in Fig. \ref{osr}.
From Fig. \ref{osr}, it is obvious that the proposed model does not have competitive results in the low-recall area, but has the best result in the high-recall area among all state-of-the-art models.
This is caused by two main factors.
First, unlike other models, the proposed model has no other structural improvements except feedforward links, making it not competitive in high-precision occasions.
Second, feedforward links play a quite significant role in relatively difficult orientation estimation tasks, making it the model with highest precision in high-recall occasions.
Higher-recall occasions involve much more small/occluded/truncated pedestrians.
In these feature-insufficient occasions, feedforward links provide useful internal information, making the proposed model more qualified for orientation estimation of those pedestrians.

\subsection{Discussions}\label{discussions}

\begin{figure}
\centering
\includegraphics[height=4cm]{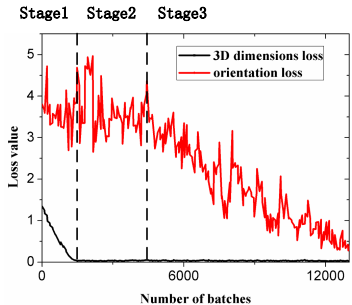}
\caption{The 3-stage convergence process of 3D dimensions and orientation loss.}
\label{respective}
\end{figure}

In the training process, the convergence processes of $loss_{dimensions}$ and $loss_{orientation}$ are shown in Fig. \ref{respective}.
The process can be divided into three stages.
At the first stage, $loss_{dimensions}$ decreases to its minimum while $loss_{orientation}$ has no sign of convergence.
This is because of the logic link from 3D dimensions output to the orientation estimator.
Orientation estimation is not accurate without precise 3D dimensions predictions.
At the second stage, $loss_{dimensions}$ has reached its minimum while $loss_{orientation}$ is still floating on a certain level.
This is because the network requires time and data to learn the logic relationship between 2D dimensions, 3D dimensions and orientation of pedestrians.
After successfully mastering the logic relationship, in the final stage, the network has its orientation loss gradually converging to its minimum.
After the 3 stages, the network reaches relatively high accuracy, and masters the logic relationship within the three characteristics.

Furthermore, we go deeper and investigate the influence on orientation estimation from 2D and 3D feedforward branch respectively.
In this experiment, the control variate method is used: the input of one branch is fixed while that of the other branch changes in a small range.
Note that in the 3D branch, the input of the feedforward link, rather than truth values of 3D dimensions, is changed.

A pedestrian with relatively fine orientation estimation accuracy is used for illustration.
Orientation and 3D dimensions errors of the prediction are roughly $0.42^\circ$ and $0.043m^2$ (mean squared error) respectively.
The characteristics of the pedestrian are listed in TABLE \ref{pedchar}.

\begin{table}[]
\centering
\caption{Characteristics of a well-predicted pedestrian for illustration.}
\begin{tabular}{cc}
\hline\hline
Characteristic & Value\\
\hline
2D height $h$ / pixel & 86\\
2D width $w$ / pixel & 33\\
3D height $h_1$ / m & 1.68\\
3D width $w_1$ / m & 0.50\\
3D length $l_1$ / m & 0.42\\
Orientation $\theta$ / $^\circ$ & -127.26\\
\hline\hline
\end{tabular}
\label{pedchar}
\end{table}

\begin{figure}
\centering
\includegraphics[height=4cm]{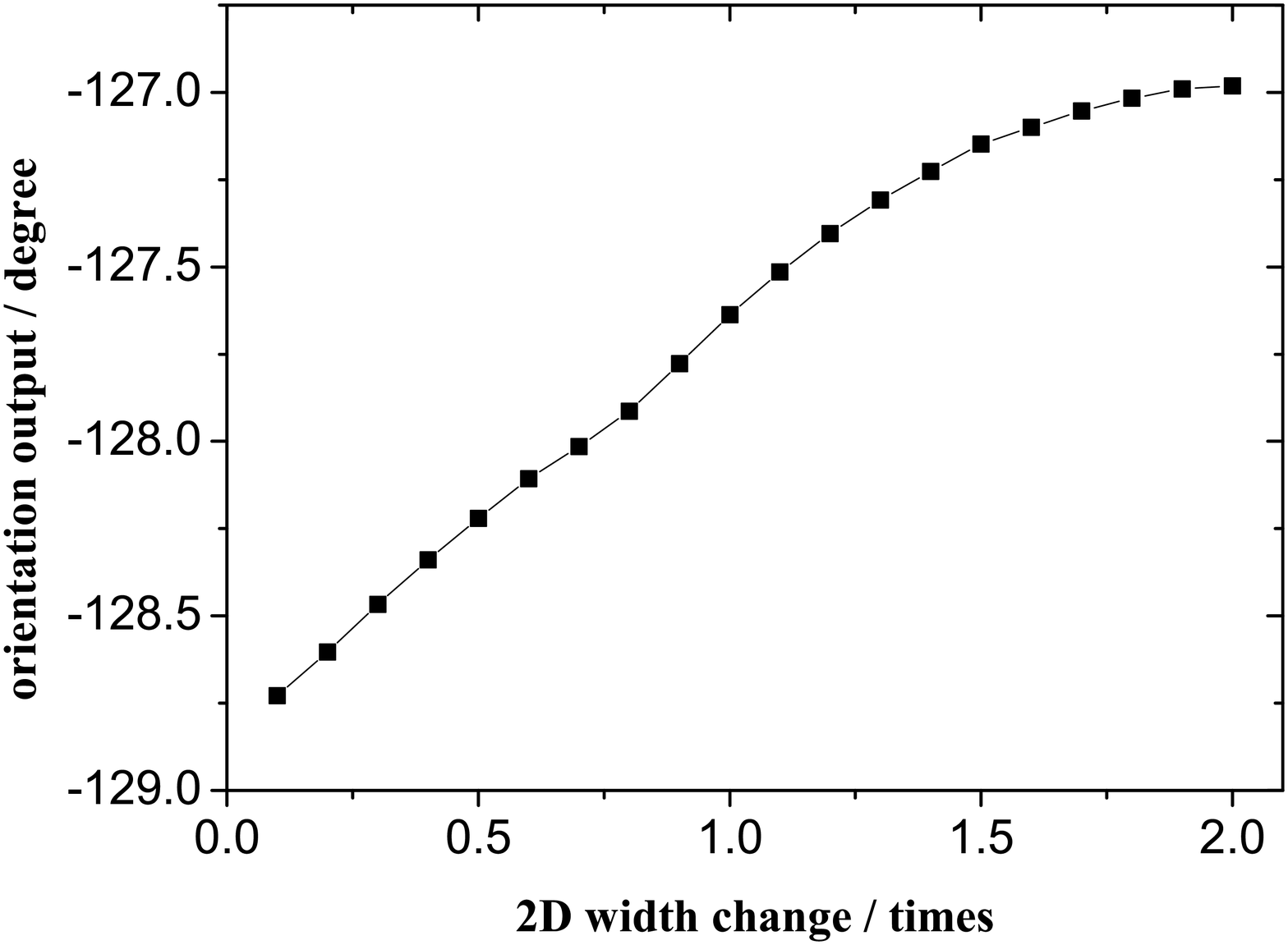}
\caption{Orientation estimations of the model with the change of the pedestrian's 2D width.}
\label{widthchange}
\end{figure}

With the change of the pedestrian's 2D width $w$ from 0.1 to 2 times the original value, orientation estimations of the model are shown in Fig. \ref{widthchange}.
Figure \ref{widthchange} shows that the model successfully learns the monotonicity of Equation \ref{eq1} with the argument $w$.
However, the amplitude of the orientation change in Fig. \ref{widthchange} is much smaller than that in Equation \ref{eq1}.
This is caused by two main factors.
First, the orientation output of the model relies largely on feature extraction and recognition of the image input (i.e. the middle link in Fig. \ref{process}). Second, there are differences between Equation \ref{eq1} and the approach of extracting and exploiting dimensions information adopted by the model, which is analyzed in Section \ref{resultplain}.

\begin{figure}
\centering
\includegraphics[height=4cm]{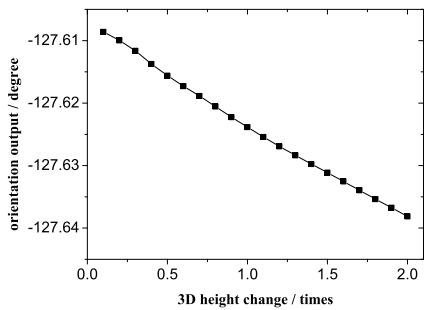}
\caption{Orientation estimations of the model with the change of the pedestrian's 3D height.}
\label{heightchange}
\end{figure}

With the change of the pedestrian's 3D height $h_1$ from 0.1 to 2 times the original value, orientation estimations of the model are shown in Fig. \ref{heightchange}.
Figure \ref{heightchange} shows that the model successfully learns the monotonicity of Equation \ref{eq1} with the argument $h_1$.
However, the amplitude of the orientation change in Fig. \ref{heightchange} is much smaller than that in Equation \ref{eq1}, with the same reasons as above.

According to Equation \ref{eq1}, a fixed 2D-3D dimensions pair corresponds up to 8 possible orientation angles.
Therefore, the proposed model yet fails to eliminate main-axes problems.
Solutions to this problem will be important future work in this field.

The proposed model shows a universal feedforward mechanism, which determines the logic sequence of the predicted variables and establishes logic links between them.
The mechanism can be adopted in various kinds of solutions dealing with multiple-prediction problems with logic relationship within.

\section{Conclusion}\label{conclusion}

In this paper, we propose a novel pedestrian orientation estimation model called FFNet.
The model makes use of the logic relationship between 2D-3D dimensions and orientation of pedestrians, and implements it by establishing logic links between them and orientation predictions.
The logic links are realized by 2D-3D feedforward branches.
Experiments show that the proposed model has a great performance increase compared with the same model without logic links and 2D-3D feedforward.
Test results on the KITTI Object Detection and Orientation Estimation Evaluation benchmark further verify our claim.
The proposed feedforward branches act as a mechanism that can be applied to a variety of solution models with multiple predictions.

\section*{Acknowledgements}

This work is supported by the National Natural Science Foundation of China (U1764264/61873165), International Chair on automated driving of ground vehicle.

\bibliographystyle{unsrt}
\bibliography{paper}

\end{document}